\documentclass[preprint,authoryear,12pt]{elsarticle}
\usepackage[doublespacing]{setspace}
\usepackage{lineno}
\usepackage{graphicx}
  \linenumbers

\usepackage{amssymb}
\journal{Pattern Recognition Letters}

\begin{document}

\begin{frontmatter}

%% Title, authors and addresses

%% use the tnoteref command within \title for footnotes;
%% use the tnotetext command for the associated footnote;
%% use the fnref command within \author or \address for footnotes;
%% use the fntext command for the associated footnote;
%% use the corref command within \author for corresponding author footnotes;
%% use the cortext command for the associated footnote;
%% use the ead command for the email address,
%% and the form \ead[url] for the home page:
%%
%% \title{Title\tnoteref{label1}}
%% \tnotetext[label1]{}
%% \author{Name\corref{cor1}\fnref{label2}}
%% \ead{email address}
%% \ead[url]{home page}
%% \fntext[label2]{}
%% \cortext[cor1]{}
%% \address{Address\fnref{label3}}
%% \fntext[label3]{}

\title{Authorship Analysis based on Data Compression}

%% use optional labels to link authors explicitly to addresses:
%% \author[label1,label2]{<author name>}
%% \address[label1]{<address>}
%% \address[label2]{<address>}

\author{Daniele Cerra, Mihai Datcu, and Peter Reinartz}

\address{German Aerospace Center (DLR), Muenchner str. 20, 82234 Wessling, Germany \\ Corresponding author's email: daniele.cerra@dlr.de, \\ phone: +49 8153 28-1496, fax: +49 8153 28-1444.}

\begin{abstract}
This paper proposes to perform authorship analysis using the Fast Compression Distance (FCD), a similarity measure based on compression with dictionaries directly extracted from the written texts. The FCD computes a similarity between two documents through an effective binary search on the intersection set between the two related dictionaries. In the reported experiments the proposed method is applied to documents which are heterogeneous in style, written in five different languages and coming from different historical periods. Results are comparable to the state of the art and outperform traditional compression-based methods. 
\end{abstract} 

\begin{keyword}
%% keywords here, in the form: keyword \sep keyword
Authorship Analysis \sep Data Compression \sep Similarity Measure
%% MSC codes here, in the form: \MSC code \sep code
%% or \MSC[2008] code \sep code (2000 is the default)
\end{keyword}

\end{frontmatter}

%%
%% Start line numbering here if you want
%%
% \linenumbers

%% main text
\section{Introduction}
\label{intro}

The task of automatically recognizing the author of a given text finds several uses in practical applications, ranging from authorship attribution to plagiarism detection, and it is a challenging one \citep{authorship_attribution}. While the structure of a document can be easily interpreted by a machine, the description of the style of each author is in general subjective, and therefore hard to derive in natural language; it is even harder  to find a description which enables a machine to automatically tell one author from the other. %The interest in these topics is confirmed by a recent special issue on authorship attribution and plagiarism detection published in \cite{special_issue_authorship}, while 
A literature review on modern authorship attribution methods, usually coming from the fields of machine learning and statistical analysis, is reported in \cite{authorship_attribution,federalist_comparison,authorship_Koppel,authorship_Grieve, Juola}. Among these, algorithms based on similarity measures such as \cite{benedetto} and \cite{authorship_burrows} are widely employed and usually assign an anonymous text to the author of the most similar document in the training data.

During the last decade, compression-based distance measures have been effectively applied to cluster texts written by different authors \citep{cilibrasi2005} and to perform plagiarism detection \citep{chen2004shared}. Such universal similarity measures, of which the most well-known is the Normalized Compression Distance (NCD), employ general compressors to estimate the amount of shared information between two objects. Similar concepts are also used by methods using runlength histograms to retrieve and classify documents \citep{runlength}. 
Experiments carried out in  \cite{forensic_authorship}  conclude that NCD-based methods for authorship analysis outperform state-of-the-art classification methodologies such as Support Vector Machines. A study on larger and more statistically meaningful datasets   shows NCD-methods to be competitive with respect to the state of the art \citep{Graaff}, while  \cite{authorship_attribution} reports that compression-based methods are effective but\texttt{} hard to use in practice as they are very slow. 

Indeed the universality of these measures comes at a price, as the compression algorithm must be run at least $n^2$ times on $n$ objects to derive a distance matrix, slowing down the analysis. Furthermore, as these methods are applied to raw data they cannot be tuned to increase their performance on a given data type. We propose then to perform these tasks using the Fast Compression Distance (FCD) recently defined in \cite{FCD}, which provides superior performances with a reduced computational complexity with respect to the NCD, and can be tuned according to the kind of data at hand. In the case of natural texts, only FCD's general settings should be adjusted according to the language of the dataset, thus keeping the desirable parameter-free approach typical of NCD.
Applications to authorship and plagiarism analysis are derived by extracting  meaningful dictionaries directly from the strings representing the data instances and matching them. The reported experiments show that improvements over traditional compression-based analysis can be dramatic, and that the FCD could become an important tool of easy usage for the automated analysis of texts, as satisfactory results are achieved skipping any parameters setting step. The only exception is an optional text preprocessing step which only needs to be set once for documents of a given language, and does not depend on the specific dataset. 

The paper is structured as follows. Section \ref{sec:1} introduces compression-based similarity measures and the FCD, which will be validated in an array of experiments reported in Section \ref{sec:2}. We conclude in Section \ref{sec:3}.

\section{Fast Compression Distance}
\label{sec:1}

Compression-based similarity measures exploit general off-the-shelf compressors to estimate the amount of information shared by any two
objects. They have been employed for clustering and classification
on diverse data types such as texts and images \citep{watanabe}, with \cite{keogh} reporting that they outperform general distance measures. The most widely known and used of such notions is the Normalized Compression Distance (NCD), defined for any two objects $x$ and $y$ as:

\begin{equation}
NCD(x,y) = \frac{C(x,y) - \min {C(x),C(y)}}{\max{C(x),C(y)}}
\end{equation}

where $C(x)$ represents the size of $x$ after being compressed
by a compressor (such as Gzip),
and $C(x, y)$ is the size of the compressed version of $x$ appended
to $y$.  If $x = y$, the NCD is approximately 0, as the full string $y$ can be described in terms of previous strings found in $x$; if $x$ and $y$ share no common information the NCD is $1+e$, where $e$ is a small quantity (usually $e < 0.1$) due to imperfections characterizing real compressors. The idea is that if $x$ and $y$ share common information
they will compress better together than separately, as the
compressor will be able to reuse recurring patterns found in
one of them to more efficiently compress the other. 
The generality of NCD allows applying it to diverse datatypes, including natural texts. Applications to authorship categorization have been presented by \cite{cilibrasi2005}, while plagiarism detection of students assignments has been succesfully carried out by \cite{chen2004shared}. 
%The NCD can be applied to any two files, and its generality allows applying it to 

A modified version of NCD based on the extraction of dictionaries has been first defined by \cite{greekdic}. The advantages of using dictionary-based methods have been then studied by \cite{FCD}, in which the authors define a Fast Compression Distance (FCD), and succesfully apply it to image analysis. The algorithm can be used for texts analysis as follows.

First of all, all special characters such as punctuation marks are removed from a string $x$, which is subsequently tokenized in a set of words $W_x$. The sequence of tokens is analysed by the encoding algorithm of the Lempel-Ziv-Welch (LZW) compressor \citep{lzw}, with the difference that words rather than characters are taken into account. The algorithm initializes the dictionary $D(x)$ with all the words $W_x$. Then the string $x$ is scanned for successively longer sequences of words in $D(x)$ until a mismatch in $D(x)$ takes place; at this
point the code for the longest pattern $p$ in the dictionary is sent
to output, and the new string ($p$ + the last word which caused
a mismatch) is added to $D(x)$. The last input word is
then used as the next starting point: in this
way, successively longer sequences of words are registered in the dictionary
and made available for subsequent encoding, with no repeated entries in $D(x)$.
An example for the encoding of the string "TO BE OR NOT TO BE OR NOT TO BE OR WHAT" after tokenization is reported
in Table \ref{lzw}. It helps to remark that the output of the simulated compression process is not of interest for us, as the only thing that will be used is the dictionary.

\begin{table}
  \centering
  \caption{LZW encoding of the tokens composing the string "TO BE OR NOT TO BE OR NOT TO BE OR WHAT". The compressor tries to substitute pattern codes referring to sequences of words which occurred previously in the text.}
  \label{LZW}
\begin{tabular}{|c|c|c|c|}
  \hline
  % after \\: \hline or \cline{col1-col2} \cline{col3-col4} ...
   Current token & Next token & Output & Added to Dictionary \\
  \hline
  Null & TO &  &  \\
  TO & BE & $TO$ & TO BE=$<1>$ \\
  BE & OR & $BE$ & BE OR=$<2>$ \\
  OR & NOT & $OR$ & OR NOT=$<3>$ \\
  NOT & TO & $NOT$ & NOT TO=$<4>$ \\
  TO BE & OR & $<1>$ & TO BE OR=$<5>$ \\
  OR NOT & TO & $<3>$ & OR NOT TO=$<6>$ \\
  TO BE OR & WHAT & $<5>$ & TO BE OR WHAT=$<7>$ \\
  WHAT & $\sharp$ & $WHAT$ &  \\
  \hline
\end{tabular}
\label{lzw}
\end{table}

The patterns contained in the dictionary $D(x)$ are then sorted in ascending alphabetical order to enable the binary search of each pattern in time $O(logN)$, where $N$ is the number of entries in $D(x)$. The dictionary is finally stored for future use: this procedure may be carried out offline and has to be performed only once for each data instance. Whenever a string $x$ is checked against a database containing $n$ dictionaries, a dictionary $D(x)$ is extracted from $x$ as described and matched against each of the $n$ dictionaries. The FCD between $x$ and an object $y$ represented by $D(y)$ is defined as:

\begin{equation}
FCD(x,y) = \frac{|D(x)| - \cap (D(x),D(y)) }{|D(x)|}
\end{equation}

where $|D(x)|$ and $|D(y)|$ are the sizes of the relative dictionaries, represented by the number of entries they contain, and $\cap(D(x),D(y))$ is the number of patterns which are found in both dictionaries. We have $FCD(x,y) = 0$ iff all patterns in $D(x)$ are contained also in $D(y)$, and $FCD(x,y) = 1$ if no single pattern is shared between the two objects. 

The FCD allows computing a compression-based distance between two objects in a faster way with respect to NCD (up to one order of magnitude), as the dictionary for each object must be extracted only once and computing the intersection between two dictionaries $D(x)$ and $D(y)$ is faster than compressing the concatenation of $x$ appended to $y$ \citep{FCD}. The FCD is also more accurate, as it overcomes drawbacks such as the limited size of the lookup tables, which are employed by real compressors for efficiency constraints: this allows exploiting all the patterns contained in a string. Furthermore, while the NCD is totally data-driven, the FCD enables a token-based analysis which allows preprocessing the data, by decomposing the objects into fragments which are semantically relevant for a given data type or application. This constitutes a great advantage in the case of plain texts, as the direct analysis of words contained in a document and their concatenations allows focusing on the relevant informational content. In plain English, this means that the matching of substrings in words which may have no semantic relation between them (e.g. `butter' and `butterfly') is prevented. Additional improvements can be made depending on the texts language. For the case of English texts, the subfix `s' can be removed from each token, while from documents in Italian it helps to remove the last vowel from each word: this avoids considering semantically different plurals and some verbal forms.

A drawback of the proposed method is that it cannot be applied effectively to very short texts. The algorithm needs to find reoccurring word sequences in order to extract dictionaries of a relevant size, which are needed in order to find patterns shared with other dictionaries. Therefore, the compression of the initial part of a string is not effective: we estimated empirically 1000 tokens or words to be a reasonable size for learning the model of a document and to be effective in its compression. 

%We estimated empirically that at least 1000 words are needed for an efficient encoding of each object. %The algorithm resembles in this the unmasking method for authorship attribution, which needs in comparison even longer texts as input \cite{unmasking}. 

\begin{figure}
%	\begin{center}
%	\includegraphics[width=5.2in]{sample_dic.jpg}
	\includegraphics[width=0.8\textwidth,natwidth=610,natheight=642]{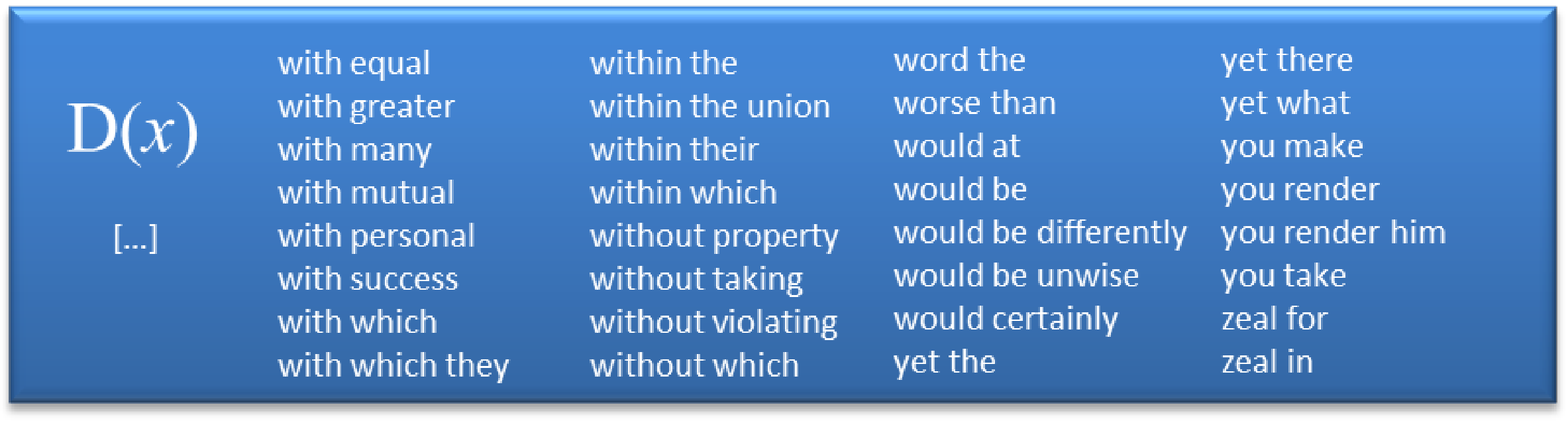}
	\caption{Subset from a dictionary $D(x)$ extracted from a sample text $x$ belonging to the Federalist dataset.} 
 	\label{sample_dic}
%	\end{center}
\end{figure}

%\paragraph{Paragraph headings} Use paragraph headings as needed.

% For tables use
%\begin{table}
% table caption is above the table
%\caption{Please write your table caption here}
%\label{tab:1}       % Give a unique label
% For LaTeX tables use
%\begin{tabular}{lll}
%\hline\noalign{\smallskip}
%first & second & third  \\
%\noalign{\smallskip}\hline\noalign{\smallskip}
%number & number & number \\
%number & number & number \\
%\noalign{\smallskip}\hline
%\end{tabular}
%\end{table}

\section{Experimental Results}
\label{sec:2}

The FCD as described in the previous section can be  effectively employed in tasks like authorship and plagiarism analysis. We report in this section experiments on four datasets written in English, Italian, and German.

\subsection{The Federalist Papers}
\label{federalist_section}

We consider a dataset of English texts known as Federalist Papers, a collection of 85 political articles written by Alexander Hamilton, James Madison and John Jay, published in 1787-88 under the anonymous pseudonym `Publius'. This corpus is particularly interesting, as Hamilton and Madison claimed later the authorship of their texts, but a number of essays (the ones numbered 49-58 and 62-63) have been claimed by both of them. This is a classical dataset employed in the early days of authorship attribution literature, as the candidate authors are well-defined and the texts are uniform in thematics \citep{authorship_attribution}. Several studies agreed on assigning the disputed works in their entirety to Madison, while Papers 18-20 have generally been found to be written jointly by Hamilton and Madison as Hamilton claimed, even though some researchers tend to attribute them to Madison alone \citep{federalist_comparison,federalist,adair}.  

\begin{figure}
%	\begin{center}
%\includegraphics[width=3.5in]{federalist.jpg}
\includegraphics[width=0.9\textwidth,natwidth=610,natheight=642]{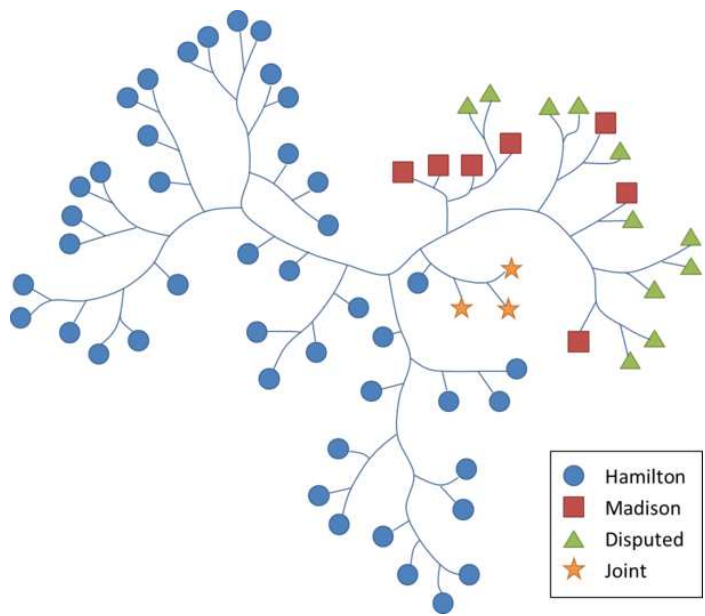}

\caption{Hierarchical clustering of the Federalist dataset, derived by a full distance matrix obtained on the basis of the FCD distance.} 
%	\end{center}
 \label{federalist_fcd}
\end{figure}

\begin{figure}
%\begin{center}
%\includegraphics[width=4in]{federalist_NCD.jpg}
\includegraphics[width=1.0\textwidth,natwidth=610,natheight=642]{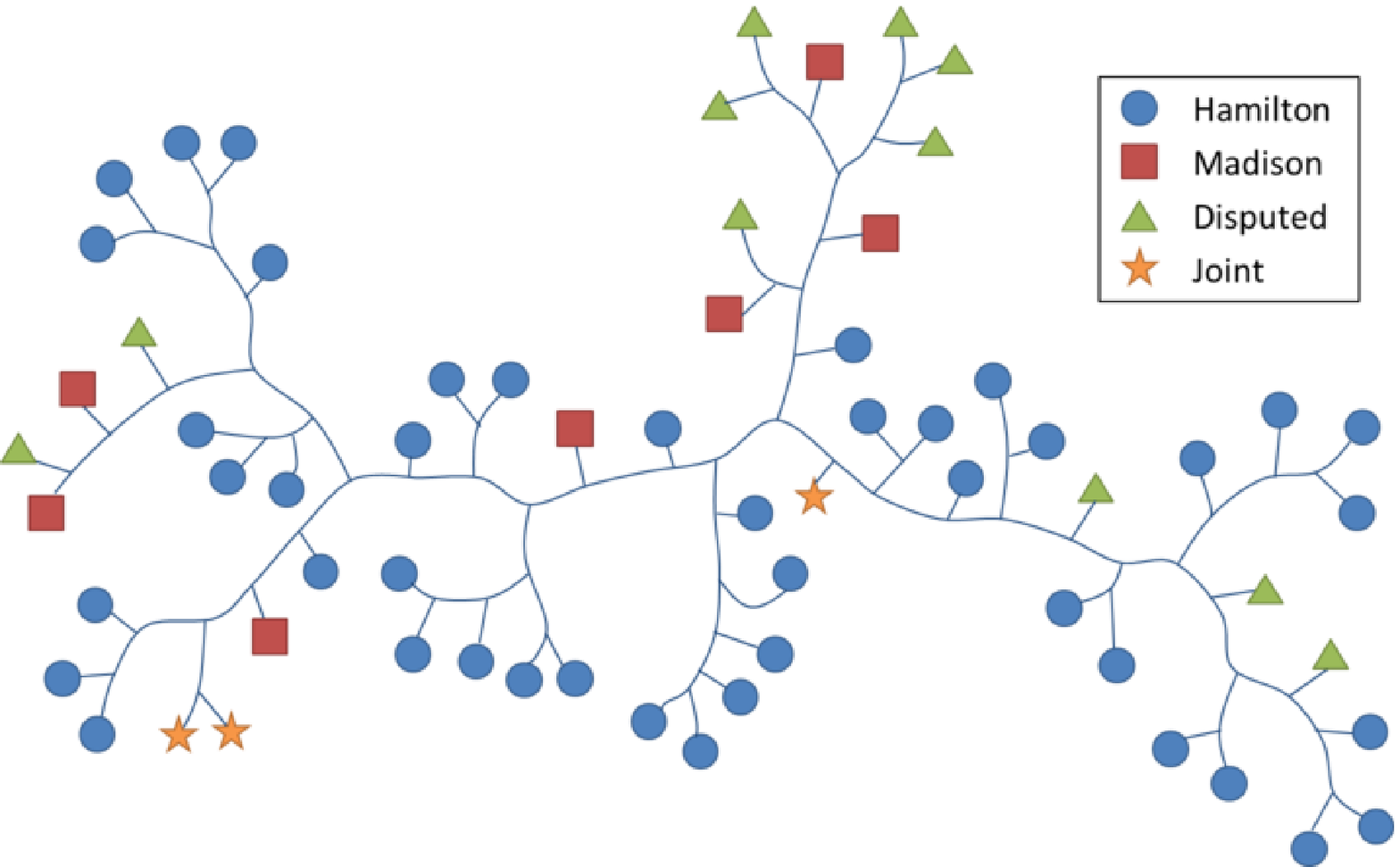}
\caption{Hierarchical clustering of the Federalist dataset obtained on the basis of the NCD distance.} 
 \label{federalist_ncd}
%	\end{center}
\end{figure}

We analyzed a dataset composed of a randomly selected number of texts of certain attribution by Hamilton and Madison, plus all the disputed and jointly written essays. We then computed a distance matrix related to the described dataset according to the FCD distance, and performed on the matrix a hierarchical clustering which is by definition unsupervised. A dendrogram (binary tree) is heuristically derived to represent the distance matrix in 2 dimensions through the application of genetic algorithms \citep{cilibrasithesis, cilibrasi2005}. Results are reported in Fig. \ref{federalist_fcd}, and have been obtained using the freely available tool CompLearn available at \cite{complearn}. Each leaf represents a text, with the documents which behave more similarly in terms of distances from all the others appearing as siblings.
The evaluation is done by visually inspecting if texts written by the same authors are correctly clustered in some branch of the tree, i.e. by checking how well the texts by the two authors can be isolated by `cutting' the tree at a convenient point. The clustering agrees with the general interpretation of the texts: all the disputed texts are clearly placed in the section of the tree containing Madison's works. Furthermore, the three jointly written works are clustered together and placed exactly between Hamilton and Madison's essays. We compare results with the hierarchical clustering derived from the distance matrix obtained on the basis of NCD distances (Fig. \ref{federalist_ncd}), run with the default blocksort compression algorithm provided by CompLearn: in this case the misplacements of the documents is evident, as disputed works are in general closer to Madison texts but are scattered throughout the tree.


\subsection{The Liber Liber dataset}

The rise of interest in compression-based methods is in part due to the concept of relative entropy as described in \cite{benedetto}, which quantifies a distance between two isolated strings relying on information theoretical notions. In this work the authors succesfully perform clustering and classification of documents: one of the considered problems is to automatically recognize the authors of a collection comprising 90 texts of 11 known Italian authors spanning the centuries XIII-XX,  available at \cite{liberliber}. Each text $x$ was used as a query against the rest of the database, its closest object $y$ minimizing the relative entropy $D(x,y)$ was retrieved, and $x$ was then assigned to the author of $y$. In the following experiment the same procedure as
\cite{benedetto} and a dataset as close as possible have been adopted, with each text $x$ assigned to the author of the text $y$ which minimizes $FCD(x,y)$. We compare our results with the ones obtained by the Common N-grams (CNG) method proposed by \cite{keselj} using the most relevant 500, 1000 and 1500 3-grams in Table \ref{authorsFCD}. The FCD finds the correct author in 97.8\% of the cases, while the best n-grams setting yields an accuracy of 90\%. For FCD only two texts, \emph{L'Asino} and \emph{Discorsi sopra la prima deca di Tito Livio}, both by Niccol\'o Machiavelli, are incorrectly
assigned respectively to Dante and Guicciardini, but these errors
may be justified: the former is a poem strongly influenced by Dante
\citep{dante}, while the latter was found similar to a collection of
critical notes on the very \emph{Discorsi} compiled by Guicciardini,
who was Machiavelli's friend \citep{machiavelli}. The N-grams-based method also assigns incorrectly Guicciardini's notes and a Dante's poem to Machiavelli, among others misclassifications. 

We also compared our results with an array of other compression-based similarity measures (Table \ref{Authorship_Comparison}): our results outperform both the Ziv-Merhav distance \citep{coutinho} and the relative entropy as described in \cite{benedetto}, while
the algorithmic Kullback-Leibler divergence \citep{entropy} obtains the same
results in a considerably higher running time. Accuracy for the NCD
method using an array of linear compressors ranged from the 93.3\%
obtained using the bzip2 compressor to the 96.6\% obtained with the
blocksort compressor. Even though
accuracies are comparable and the dataset may be small to be
statistically meaningful, another advantage of FCD over NCD is the
decrease in computational complexity. While for NCD it took 202
seconds to build a distance matrix for the 90 pre-formatted texts
using the zlib compressor (with no appreciable variation when using
other compressors), just 35 seconds were needed on the same machine
for the FCD: 10 to extract the dictionaries and the rest to build
the full distance matrix.

\begin{table}
\centering
\caption{Classification results on the Liber Liber dataset. Each text from the 11 authors is used to query the database, and it is considered to be written by the author of the most similar retrieved work. The authors' full names: Dante Alighieri, Gabriele D'Annunzio, Grazia Deledda, Antonio Fogazzaro, Francesco Guicciardini, Niccol\'o Machiavelli, Alessandro Manzoni, Luigi Pirandello, Emilio Salgari, Italo Svevo, Giovanni Verga. The CNG method has been tested using the reported amounts of n-grams.}
\begin{tabular}{|l|c|c|c|c|c|}
  \hline
  % after \\: \hline or \cline{col1-col2} \cline{col3-col4} ...
Author & Texts & FCD & CNG-500 & CNG-1000 & CNG-1500 \\
\noalign{\smallskip}\hline\noalign{\smallskip}
Dante Alighieri & 8 & 8 & 6 & 5 & 7\\
  D'Annunzio & 4 & 4 & 4 & 3 & 4 \\
  Deledda & 15 & 15 & 15 & 15 & 14 \\
  Fogazzaro & 5 & 5 & 4 & 5 & 5 \\
  Guicciardini & 6 & 6 & 5 & 5 & 5 \\
  Machiavelli & 12 & 10 & 8 & 10 & 9\\
  Manzoni & 4 & 4 & 4  & 4 & 4 \\
  Pirandello & 11 & 11 & 5 & 10 & 8 \\
  Salgari & 11 & 11 & 10 & 10 & 9 \\
  Svevo & 5 & 5 & 4 & 5 & 5 \\
  Verga & 9 & 9 & 6 & 9 & 8 \\
  \hline
  Total & 90 & 88 & 71 & 81 & 78 \\
  Accuracy (\%) & 100 & 97.8 & 78.9  & 90 & 86.7 \\
  \hline
\end{tabular}
\label{authorsFCD}
\end{table}

\begin{table}
\centering
\caption{Accuracy and running time for different compression-based methods applied to the Liber Liber dataset.}
\begin{tabular}{|l|c|c|}
  \hline
  % after \\: \hline or \cline{col1-col2} \cline{col3-col4} ...

Method & Accuracy (\%) & Running Time (sec) \\
 \noalign{\smallskip}\hline\noalign{\smallskip}
FCD & 97.8 & 35 \\
Relative Entropy & 95.4 & NA \\
Ziv-Merhav & 95.4 & NA \\
NCD (zlib) & 94.4 & 202 \\
NCD (bzip2) & 93.3 & 198 \\
NCD (blocksort) & 96.7 & 208 \\
Algorithmic KL & 97.8 & 450 \\
  \hline
\end{tabular}
\label{Authorship_Comparison}
\end{table}

%\begin{figure}
%\centering 
%\includegraphics[width=3.5in]{fig/PrecRec.png}
% \caption{Precision vs. Recall graph for the Liber Liber dataset for FCD and NCD using different compressors.}
% \label{prec_rec}
%\end{figure}

%It would be interesting to consider how well a content-based retrieval system based on these notions would perform. In a classical Query By Example (QBE) architecture the user presents to the system a sample object and retrieves all instances in a database which are similar, according to given criteria \cite{Smeulders}. We simulate a QBE system by using as query in turn every object in the dataset: the average result of all the queries is depicted in the Precision vs. Recall curve reported in Fig \ref{prec_rec}. The Precision related to a query is defined as the number of relevant documents retrieved divided by the total number of documents
%retrieved, while Recall is defined as the number of relevant documents retrieved divided by the total number of existing relevant documents \cite{baeza-yates}. 

\subsection{The PAN Benchmark Dataset}

%In order to test the method on a recent state-of-the-art benchmark dataset for authorship attribution,
We tested our algorithm on datasets from the two most recent \cite{pan} competitions, which provide benchmark datasets for authorship attribution. From PAN 2013 we selected the author identification task described in \cite{PAN13results}. In this task 349 training texts are provided, divided in 85 problems out of which 30 are in English, 30 in Greek and 25 in Spanish. For each set of documents written by a single author it must be determined if a questioned document was written by the same author or not. Each text is approximately 1000 words long, which is close to our empirical estimation of the minimum size for FCD to find relevant patterns in a data instance (Section \ref{sec:1}). For each problem, we consider an unknown text to be written by the same author of a given set of documents if the average FCD distance to the latter is smaller than the mean distance from all documents of a given language. Compared to the performance of the 18 methods reported in \cite{PAN13results}, the FCD finds the correct solution in $72.9\%$ of the cases and yields the second best results, ranking first for the set of English problems and fifth for both the Greek and Spanish sets (Table \ref{PAN13}), outperforming among others two compression-based and several n-grams-based methods. It must be stressed that the FCD took approximately 38 seconds to process the whole dataset, while the imposters method by \cite{pan13_seidman}, which ranked first in the competition for all problems excluded the ones in Spanish, took more than 18 hours. Furthermore, the latter method requires the setting of a threshold, while the FCD skips this step. On the other hand, the contest participants had only a small subset of the available ground truth to test their algorithms.

\begin{table}
\centering
\caption{Author identification task of the CLEF PAN 2013 dataset. The dataset contains 349 training texts plus 85 test documents of questioned authorship, with problems given in English, Greek and Spanish. The table reports how the FCD ranks compared to 18 participants to the PAN 2013 contest. The first ranked submission for each problem is reported as `Best PAN'. }
\begin{tabular}{|c|c|c|c|c|}
  \hline
  % after \\: \hline or \cline{col1-col2} \cline{col3-col4} ...

Task & FCD & Best PAN & Rank \\
\noalign{\smallskip}\hline\noalign{\smallskip}

 Overall     & 72.9 \% & 75.3 \% & 2 \\
English     & 83 \% & 80 \% & 1 \\
Greek    & 63 \% & 83 \% & 5 \\
Spanish       & 72 \% & 84 \% & 5 \\
%Running Time & 35s & 65477s & 6 \\
  \hline
\end{tabular}
\label{PAN13}
\end{table}

\begin{figure}
\includegraphics[width=1.0\textwidth,natwidth=610,natheight=642]{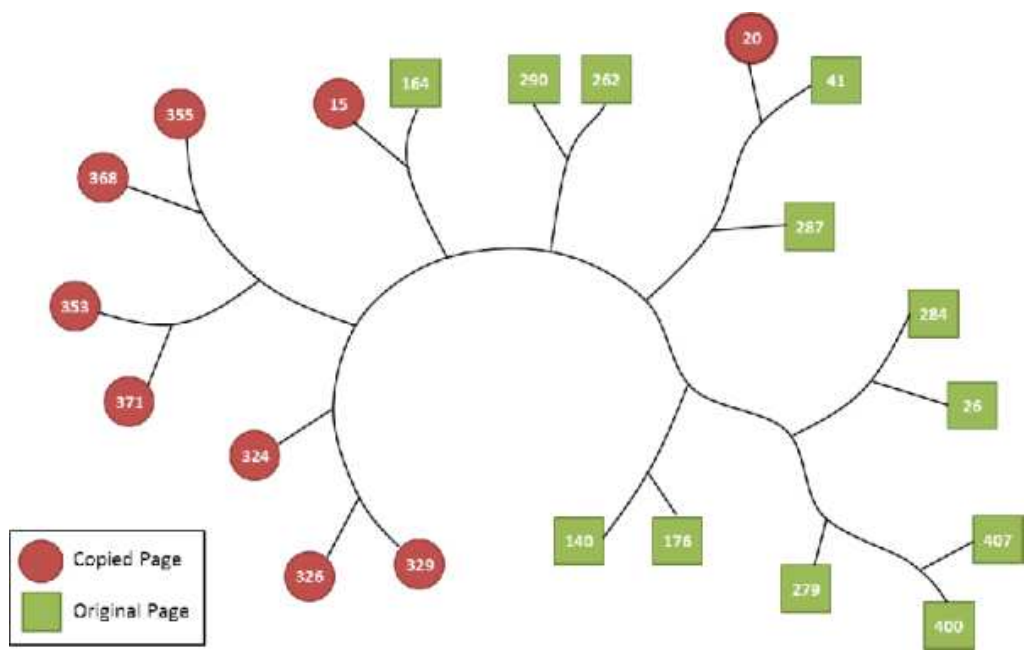}
\caption{Hierarchical clustering of pages extracted from Guttenberg PhD thesis.}
\label{guttenberg_fcd}
\end{figure}

%We were not able to process all the available datasets, as open-class problems could not be analysed as the simple classification algorithm adopted does not allow a rejection class.

We tested FCD also on the largest closed-class classification problem (task I) from the 2012 PAN competition: open-class problems were not considered as the simple classification algorithm adopted does not allow a rejection class. Using a corpus of 14 test and 28 training texts belonging to 14 different authors, the FCD (using a simple nearest neighbour classification criterion) assigns correctly 12 out of 14 documents to their correct authors. Out of the 25 which took part to the competition, only 4 methods submitted by three groups  \citep{pan1,pan2,pan3} %, \citep{pan2}, and \citep{pan3}
outperformed our method (all of them with 13 documents correctly recognized). As a comparison, the NCD and trigrams-based CNG (using the most meaningful 1000 trigrams per document, as this setting yields the best results in Table \ref{authorsFCD}) assigned 2 and 9 documents out of 14 to the correct author, respectively. The results in Tables \ref{PAN13} and \ref{PAN12} are encouraging, specially if we consider that the FCD is a general method which is not specific for the described tasks.

\begin{table}
\centering
\caption{Classification results on task I of the CLEF PAN 2012 dataset. The dataset contains 28 texts belonging to 14 different authors for training and 14 for testing. The best results obtained in the PAN 2012 contest are reported as `Best'. }
\begin{tabular}{|c|c|c|c|c|}
  \hline
  % after \\: \hline or \cline{col1-col2} \cline{col3-col4} ...

Method & FCD & NCD & CNG & Best\\
\noalign{\smallskip}\hline\noalign{\smallskip}
 Correct (out of 14)    & 12 & 2 & 9 & 13\\
  \hline
\end{tabular}
\label{PAN12}
\end{table}

\subsection{The Guttenberg Case}

In February 2011, evidence was made public that the former German minister Karl-Theodor zu Guttenberg had violated the academic code by copying several passages of his PhD thesis without properly referencing them. This eventually led to Guttenberg losing his PhD title, resigning from being minister, and being nicknamed Baron Cut-and-Paste, Zu Copyberg and Zu Googleberg by the German media \citep{guttenberg_news}. Evidence of the plagiarism and a detailed list of the copied sections and of the different sources used by the minister is available at \cite{guttenplag}.

We selected randomly two sets of pages from this controversial dissertation, with the first containing plagiarism instances, and the second material originally written by the ex-minister. Then we performed an unsupervised hierarchical clustering on the distance matrix derived from FCD distances as described in Section \ref{federalist_section}. First attempts made by analyzing single pages failed at separating the original pages in a satisfactory way, as the compressor needs a reasonable amount of data to be able to correctly identify shared patterns between the texts. We selected then two-pages long excerpts from the thesis, with the resulting clustering reported in Fig. \ref{guttenberg_fcd} showing a good separation of the texts containing plagiarism instances (in red in the picture). The only confusion comes from pages starting at 41 with pages starting at 20, in the bottom-left part of the clustering. This is justified by the fact that page 41 refers to the works of Loewenstein, who happens to be the same author from which part of page 20 was plagiarized \citep{Loewenstein}. Therefore, the system considers page 20 to be similar to the original style of the author at page 41.

Even though the described procedure is not able to detect plagiarism, it can find excerpts in a text which are similar to a given one. If instances of plagiarized text can be identified, objects close to them in the hierarchical clustering will be characterized by a similar style: therefore, this tool could be helpful in identifying texts which are most likely to have been copied from similar sources.

\section{Conclusions}
\label{sec:3}

This paper evaluates the performance of compression-based similarity measures on authorship and plagiarism analysis on natural texts. Instead of the well-known Normalized Compression Distance (NCD), we propose using the dictionary-based Fast Compression Distance (FCD), which decomposes the texts in sets of reoccurring combinations of words captured in a dictionary, which describe the text regularities, and are compared to estimate the shared information between any two documents. The reported experiments show the universality and adaptability of these methods, which can be applied without altering the general workflow to documents written in English, Italian, Greek, Spanish and German. The main advantage of the FCD with respect to traditional compression-based methods, apart from the reduced computational complexity, is that it yields more accurate results. We can justify this with two remarks: firstly, the FCD should be more robust since it performs a word-based analysis, focusing exclusively on meaningful patterns which better capture the information contained in the documents; secondly, the use of a full dictionary allows discarding any limitation that real compressors have concerning the size of buffers and lookup tables employed, being the size of the dictionaries bounded only by the number of relevant patterns contained in the objects. At the same time, the data-driven approach typical of NCD is maintained. This allows keeping an objective, parameter-free workflow for all the problems considered in the applications section, in which promising results are presented on collections of texts in Italian, English, and German.

\end{document}